# On the Complexity of Dark Chinese Chess

Cong Wang
School of Computer Science and Engineering
Wuhan Institute of Technology
Wuhan, China, 430073
e-mail: xyfcw@qq.com

Tongwei Lu
School of Computer Science and Engineering
Wuhan Institute of Technology
Wuhan, China, 430073
e-mail: lutongwei@wit.edu.cn

*Abstract*—This paper provides a complexity analysis for the game of dark Chinese chess (a.k.a. "JieQi") , a variation of Chinese chess. Dark Chinese chess combines some of the most complicated aspects of board and card games, such as long-term strategy or planning, large state space, stochastic, and imperfect-information, which make it closer to the real world decision-making problem and pose great challenges to game AI. Here we design a self-play program to calculate the game tree complexity and average information set size of the game, and propose an algorithm to calculate the number of information sets.

## I. INTRODUCTION

In recent years, artificial intelligence (AI) has been competing with human beings in the field of games, and has made great progress in perfect-information games like Go [1, 2]. In some sense, two-player zero-sum perfect-information games have been solved by the learning algorithm AlphaZero [3]. Therefore, the focus of recent research turns to imperfect-information games, where the players compete with others in a partially observable environment [4-8].

Dark Chinese chess is a variation of Chinese chess, which is a stochastic and imperfect-information game. Board games such as Chess, Go and Chinese Chess all have variants of imperfect-information [9-13]. For example, Chinese dark chess is a variant of Chinese chess. In the beginning of the game, all pieces are randomly placed on the board with piece icons facing down and can only be moved when they are turned over [13]. Dark Chinese chess can be viewed as an upgraded version of Chinese dark chess, increasing the size of the state space while preserving imperfect-information.

In chess games, general algorithms such as AlphaZero can not be applied to dark Chinese chess, because the algorithm does not have the ability to deal with random, imperfect information states. In the field of imperfect-information games, counterfactual regret minimization (CFR) algorithm [14] and its variants combined with some abstract technologies have made a breakthrough in Texas Hold'em poker [4-6]. But these algorithms can not be directly applied to dark Chinese chess. First, dark Chinese chess not only contains uncertain factors, but also contains the huge action space common in board games. Second, different from card games, the value of the pieces in the game is related to the position, and the value of the pieces in different positions varies greatly. The above features of the game make the abstract technologies difficult to use.

As abstractions of the real world, games are an ideal platform for testing AI technology. As the epitome of war, chess games have a special position. Dark Chinese Chess can be regarded as the epitome of modern war and the duel between the two sides in an uncertain environment. Therefore, studying such issues is crucial to reality.

In this work, we design a self-play program and propose a algorithm to measure the complexity of dark Chinese chess. The results show that while it maintains the complexity of chess strategy, it adds more uncertainty than card games. The characteristics of dark Chinese chess pose challenges to game AI.

## II. DARK CHINESE CHESS

Dark Chinese chess is a variation of Chinese chess (a.k.a. "JieQi"), which has increased in popularity in China and Vietnam with tournaments held every year.

The pieces and board in dark Chinese chess are the same as those in Chinese chess. In the beginning of the game, all pieces except Kings will be flipped down and placed on the board out of order as shown in Fig. 1. These pieces are now called dark pieces. The dark chess moves according to the position, in which position according to the rules of Chinese chess pieces in this position to go. For example, the dark pieces in the original position of the pawn in Chinese chess must move as pawns (i.e. only one step forward). They will be flipped up randomly in the first move of them. The dark piece being flipped is called a revealed piece. Then, the revealed pieces move according to the rules of Chinese chess. Note that different from Chinese chess, the Guard and Minister can cross the river in dark Chinese chess.

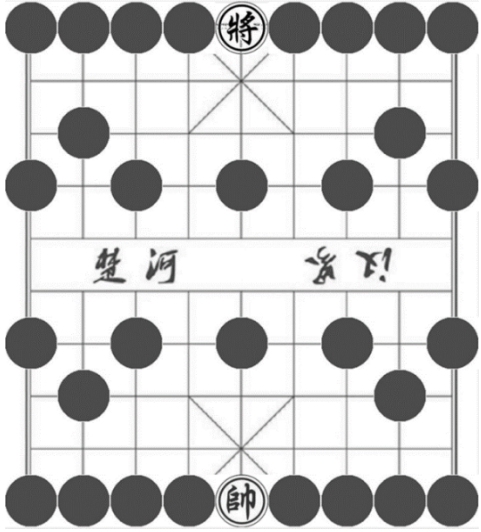

Fig. 1. The beginning state of the game.

Another thing to note is that if a player captures an opponent's dark piece, the opponent will learn that he or she lost a dark piece, but does not know what it is. This means that the information known to both players goes from symmetric to asymmetric, which not only increases the fun of the game, but also obviously makes the game more difficult.

## III. GAME COMPLEXITY

In general, the complexity of a game can be measured by state space complexity (SSC) and game tree complexity (GTC) [17]. Game Tree Complexity represents the number of all different game paths for a game. The State Space Complexity of a game is the total number of states that can be reached in accordance with the rules since the initial state of the game. However, in imperfect-information games, a reasonable game strategy should be based on the information set rather than the game state, due to the information is imperfect and asymmetric. Accordingly, when we measure the size of imperfect information games, we should measure the number of information sets, not the number of the game's states.

### A. Game Tree Complexity

Obviously, the exact value of game tree complexity is a difficult to calculate. The common method is to estimate its reasonable lower bound We choose to use Shannon's method of calculating the size of the chess game tree [16] to estimate the game tree complexity of dark Chinese chess. The approach is illustrated by (1), where $b$ represents a branching factor, the average number of legal moves per turn by the player, and $p$ represents the average length of the game. In these calculations, the length of the game is measured in plies. A ply is a single turn completed by a single player, which is significantly different from a round.

$$GTC \geq b^p \qquad (1)$$

### B. Number of Information Sets

We use the number of information sets to measure the size of imperfect-information games. An information set is a game state that is indistinguishable from a particular player's perspective. For example, in the beginning state of dark Chinese chess, we cannot distinguish what the dark piece is, so all possible states will be grouped into the same information set. Therefore, in a perfect-information game, the size of the information set is always 1. The size of a game can be measured by the number of information sets the player is likely to encounter in that game. The size of a finite game is a fixed number, but often difficult to calculate. In perfect-information games, the number of moves varies for different situations, and the length of each game varies.

### C. Size of Information Set

Another term used to measure imperfect-information games is the average size of the information set, that is, the average number of indistinguishable game states in the information set.

It is easy to see that the average size of the information set reflects the amount of information hidden behind each situation in the game. Obviously, the larger the average size of the information set, the more unknown information it contains, and therefore the more difficult the decision becomes. In fact, the average size of the information set directly affects the feasibility of the search algorithm: when the opponent's hidden state is very much, the traditional search algorithm is basically unable to start. Therefore, the average size of the information set can also be used as a measure of the difficulty of a game. This item illustrates how difficult it is to evaluate the state of a given set of information. The number of possible states can be measured in considering the number of specific states that a game can have in the hidden information under a given information set. In dark Chinese chess, this involves taking into account the number of unknown pieces, and counting the number of all possible permutations of pieces on the board. For example, the information set size of the beginning state of the game is $10^{17}$ roughly.

## IV. EXPERIMENT

### A. Self-Play

We use the Monte Carlo method to calculate game tree complexity and mean information set size of dark Chinese chess through a large number of samples. Specifically, we designed a self-play program, in each ply of game, the program computers and stores all legal moves and choose one from it randomly to play game until it is over. The game ends under three conditions. First, the game ends when the king of one player is captured. Second, when the "meet the marshals" occurs, the player who makes the "meet the marshals" situation loses. Finally, the game ends in a draw when both players do not capture a piece within 40 plies. We played 10000 self-play games, saved the length of the game and calculated the average legal moves and mean information set size in each game. Then we calculated the average size of the data, which means adding the data of each round and dividing it by the number of rounds of the game. The results are shown in Fig.2.

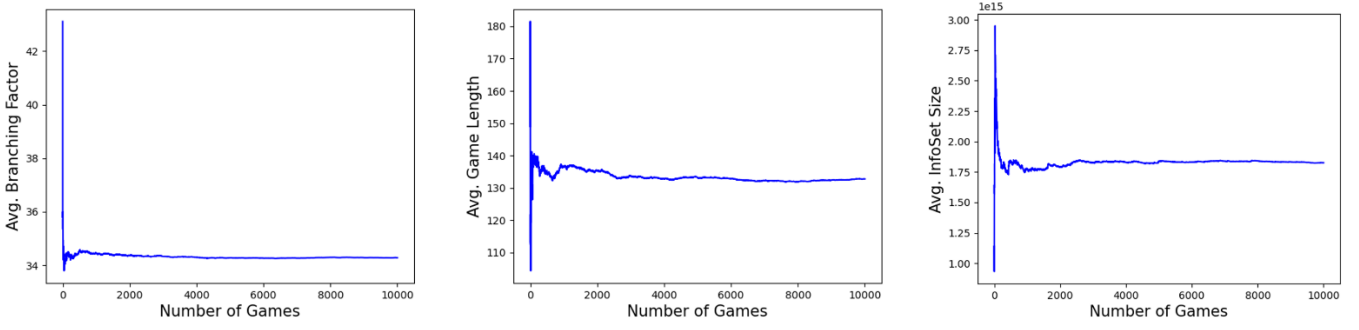

Fig. 2. The average size of branching factor, game length and information set with the increase of game rounds.

As can be seen from Fig. 2, the data fluctuated a little in the early stage and then stabilized. Results show that the branching factor is about 34, the game length is about 133 and the average of information set is $10^{15}$. Therefore, it is easy to calculate that the game tree complexity of the game is $10^{205}$.

*B. Calculate Number of Information Sets*

We calculate the number of information sets from the perspective of the red player because of the symmetry of the board and the pieces. The kings exists in all statistical states, so we can only consider the 30 pieces left, and only consider the 88 positions on the 9×10 chessboard (the kings need to occupy two positions). For the convenience of calculation, we assume that the 15 pieces on each side are different.

We first determine the number of red and black pieces, bright and dark pieces inside and outside the chessboard, and then calculate the number of information sets in this state. In order to avoid repeated calculation, we first use the three-dimensional array $R$ to store all the combined numbers when the number of red chess on the chessboard, the number of red dark chess in the chessboard and the number of red dark chess outside the chessboard are determined. From the perspective of the red side, the captured black chess, whether bright or dark, does not affect the red side's judgment of the black chess in the chessboard. Therefore, the two-dimensional array $B$ is used to store all the combined numbers when the number of black chess and the number of dark particles in the chessboard are determined. Also note that when all the red chess pieces on the chessboard are bright, the bright and dark of the red chess outside the chessboard need not be considered. See Algorithm 1 for the specific implementation.

Finally, the number of dark Chinese chess information sets is about $10^{57}$.

Algorithm 1 Calculate Number of Information Sets
Output: $num$
1: Initialize $R[i][j][k] \leftarrow C_{15}^i * C_i^j * C_{15-i}^k$, $B[i][j] \leftarrow C_{15}^i * C_i^j$ and $num \leftarrow 0$
2: For $R_{on} = 0, 1, \ldots$ until 15 do
3:    For $B_{on} = 0, 1, \ldots$ until 15 do
4:       $A_{on} \leftarrow R_{on} + B_{on}$
5:       For $B_{dk} = 0, 1, \ldots$ until $B_{on}$ do
6:          For $R_{odk} = 0, 1, \ldots$ until $R_{on}$ do
7:             $A_{dk} \leftarrow R_{odk} + B_{dk}$
8:             If $R_{dk} = 0$ then
9:                $B_1 \leftarrow C_{15}^{R_{on}} * B[B_{on}][B_{dk}]$
10:               $t_1 \leftarrow B_1 * B_t[A_{on}][A_{dk}] * C_{15}^{B_{dk}} * C_{15}^{R_{odk}}$
11:               $num \leftarrow num + t_1$
12:             Else
13:               For $R_{fdk} = 0, 1, \ldots$ until $15 - R_{on}$ do
14:                  $R_{dk} \leftarrow R_{odk} + R_{fdk}$
15:                  $B_2 \leftarrow R[R_{on}][R_{odk}][R_{fdk}] * B[B_{on}][B_{dk}]$
16:                  $t_2 \leftarrow B_2 * B_t[A_{on}][A_{dk}] * C_{15}^{B_{dk}} * C_{15}^{R_{odk}}$
17:                  $num \leftarrow num + t_2$
18:               End for
19:             End if
20:             End for
21:          End for
22:       End for
23:    End for

*C. Comparison to Other Games*

Table 1 Compared to the game tree complexity of other games

| Game | Branching factor | Average game length | Game-tree complexity(as log to base 10) |
|---|---|---|---|
| Gomoku(15*15) | 210 | 30 | 70 |
| Chess | 35 | 70 | 123 |
| Chinese chess | 38 | 95 | 150 |
| Dark Chinese chess | 35 | 133 | 205 |
| Go | 250 | 150 | 360 |

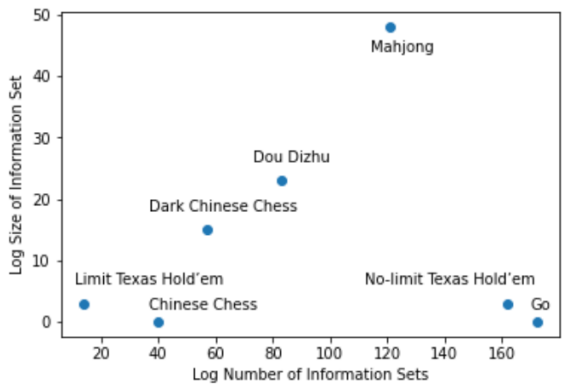

Fig. 3. Comparison of the size and number of information sets.

We can compare the complexity of dark Chinese chess with some popular games, and the results are shown in Fig. 3. and Table. 1. The figure and table clearly show that, on the basis of Chinese chess, the difficulty of the game is greatly increased simply by flipping down all pieces except Kings. This shows that dark Chinese chess will bring great challenges to the research of game AI.

## V. CONCLUSION

This paper presents a complexity analysis for dark Chinese chess. The experimental results show that dark Chinese chess has some unique characteristics that are not found in other games. Compared with many games with imperfect information, such as card games and common wargame games, the distribution of chess players' own pieces is also unknown. It retains the original strategic complexity of chess games and adds the uncertainty of card games, which brings great challenges to the research of game AI. The complexity and unique characteristics of dark Chinese chess provide a good experimental platform for the research of such problems.

This paper is the first attempt to study this interesting and complicated game. We hope it can promote future researches for dark Chinese chess.

## ACKNOWLEDGMENT

We would like to thank Xiaoyong Mao, our program is based on his work.